\documentclass[conference]{IEEEtran}
\IEEEoverridecommandlockouts
\usepackage{cite}
\usepackage{amsmath,amssymb,amsfonts}
\usepackage{algorithmic}
\usepackage{graphicx}
\usepackage{textcomp}
\usepackage{xcolor}
\usepackage{caption}
\usepackage[utf8]{inputenc}
\usepackage{subcaption}
\usepackage{enumerate}
\usepackage{enumitem}
\usepackage{booktabs}
\usepackage{flushend}

\def\BibTeX{{\rm B\kern-.05em{\sc i\kern-.025em b}\kern-.08em
    T\kern-.1667em\lower.7ex\hbox{E}\kern-.125emX}}
\begin{document}

\title{Exploring the limits of multifunctionality across different reservoir computers\\ \thanks{$^{*}$\textit{Corresponding author} \\ \vspace{0.1cm} AF is funded by the Irish Research Council Enterprise Partnership Scheme (Grant No. EPSPG/2017/301).}}


\author{
\IEEEauthorblockN{Andrew Flynn$^{*}$}
\IEEEauthorblockA{\textit{School of Mathematical Sciences} \\
\textit{University College Cork}\\
Cork, Ireland \\
andrew\_flynn@umail.ucc.ie}
\and
\IEEEauthorblockN{Oliver Heilmann}
\IEEEauthorblockA{\textit{Department of Physics} \\
\textit{Ludwig-Maximilians University} \\
Munich, Germany \\
o.heilmann@campus.lmu.de
}
\and
\IEEEauthorblockN{Daniel Köglmayr}
\IEEEauthorblockA{\textit{Department of Physics} \\
\textit{Ludwig-Maximilians University} \\
Munich, Germany \\
d.koeglmayr@physik.uni-muenchen.de
}
\and
\IEEEauthorblockN{Vassilios A. Tsachouridis}
\IEEEauthorblockA{\textit{Collins Aerospace} \\
Cork, Ireland \\
vassilios.tsachouridis@collins.com}
\and
\IEEEauthorblockN{Christoph Räth}
\IEEEauthorblockA{\textit{Institut für KI-Sicherheit} \\
\textit{Deutsches Zentrum für Luft-und Raumfahrt}\\
Wessling, Germany \\
christoph.raeth@dlr.de}
\and
\IEEEauthorblockN{Andreas Amann}
\IEEEauthorblockA{\textit{School of Mathematical Sciences} \\
\textit{University College Cork}\\
Cork, Ireland \\
a.amann@ucc.ie}
}

\maketitle

\IEEEpubidadjcol

\begin{abstract}

Multifunctional neural networks are capable of performing more than one task without changing any network connections. In this paper we explore the performance of a continuous-time, leaky-integrator, and next-generation `reservoir computer' (RC), when trained on tasks which test the limits of multifunctionality. In the first task we train each RC to reconstruct a coexistence of chaotic attractors from different dynamical systems. By moving the data describing these attractors closer together, we find that the extent to which each RC can reconstruct both attractors diminishes as they begin to overlap in state space. In order to provide a greater understanding of this inhibiting effect, in the second task we train each RC to reconstruct a coexistence of two circular orbits which differ only in the direction of rotation. We examine the critical effects that certain parameters can have in each RC to achieve multifunctionality in this extreme case of completely overlapping training data.

\end{abstract}

\begin{IEEEkeywords}
Reservoir Computing; Multifunctionality; Floquet analysis
\end{IEEEkeywords}

\section{Introduction}\label{sec:Intro}






Advancements in machine learning oftentimes arise from a `two-way street' between neuroscientific observation and mathematical representation. By following this approach, `multifunctional reservoir computing' \cite{flynn2021multifunctionality} has emerged as a means of training an artificial neural network to perform more than one task without the need to switch between different network configurations unlike other approaches such as modular neural networks \cite{WolpertKawato98}, or conceptors \cite{jaeger2017conceptors}.

`Multifunctionality' describes the ability of a single neural network to perform a multitude of mutually exclusive tasks, in other words, they possess a form of multistability. There are many examples of multifunctional neural networks in nature, for further reading we suggest \cite{briggman08multifunctional}.


The types of multistabilities that multifunctional reservoir computers (RCs) have been trained to produce range from reconstructing the dynamics of chaotic attractors which already coexist, to creating a coexistence of chaotic attractors from two different dynamical systems \cite{flynn2021multifunctionality,herteux2020Symm}. Furthermore, in \cite{flynn2021symmetry}, it was shown that even when there are overlapping regions between these attractors, a RC can also achieve multifunctionality. However, this required increasing the spectral radius of the RCs internal connections from the case where there was no overlap to achieve multifunctionality.

As no multistable autonomous dynamical system will ordinarily have coexisting attractors which share common regions of state space, this brings into question what are the essential conditions for a RC to achieve multifunctionality in this scenario. In order to gain a greater understanding of the limits of multifunctionality in the case of overlapping training data, in this paper we examine the critical effects that certain parameters can have in the training of different types of RCs. Furthermore, by exploring these limitations we are able to come closer to establishing the extent of the dynamical functionality a given RC can reconstruct.

In our numerical experiments we investigate the behaviour of three RC setups, a continuous-time RC (CT-RC) \cite{LuHuntOtt18RC}, a leaky-integrator RC (LI-RC) \cite{jaeger2007LeakyInt}, and the recently introduced `next generation' RC (NG-RC) \cite{gauthier2021ngrc}. Each of these RCs are trained using a ridge regression approach.


These experiments consist of training a RC to achieve multifunctionality by reconstructing a coexistence of chaotic attractors from the Lorenz and Halvorsen systems like the example used in \cite{herteux2020Symm}. In this task we examine the performance of each RC setup as the data describing each attractor are moved closer together. For a given set of training parameters we find that each RC can reconstruct both attractors reasonably well when they are well separated in state space. However, for the same set of training parameters, the closer the attractors are to one another, the poorer each RC performs.

To place further emphasis on the influence of these training parameters, in the second task we study an extreme case of overlapping training data. In this experiment each RC is trained to reconstruct a coexistence of trajectories on two completely overlapping circular orbits rotating in opposite directions. In order for the CT-RC and LI-RC to achieve multifunctionality in this scenario, these RCs are required to convert this overlapping driving input data into attractors with two distinct basins of attraction that can be projected to resemble the desired coexistence. While for the NG-RC, we find that it achieves multifunctionality in a different way due to its design. Instead of generating trajectories on limit cycles, the trained NG-RC is a linear system which produces centers.

Overall, our results indicate that the LI-RC outperforms the CT and NG-RC on these tasks. Despite arguing that memory is a key element for the RCs to achieve multifunctionality in the case of overlapping training data, we find that a commonly used `memory capacity' metric \cite{jaeger2002short} is insufficient in distinguishing whether different random realisations of the RCs can give rise to multifunctionality. Instead, we find that a Floquet analysis provides a more rigorous assessment.

The rest of the paper is organised as follows: In Sec.\,\ref{sec:MFRC} we outline the steps involved in training each RC setup to achieve multifunctionality. In Sec.\,\ref{sec:BenchmarkTests} we describe the numerical experiments in greater detail and the analysis tools used to assess the performance each RC. In Sec.\,\ref{sec:Results} we discuss our results and in Sec.\,\ref{sec:Conc} we provide some concluding remarks.

\section{Multifunctional Reservoir Computing}\label{sec:MFRC}








Since the introduction of echo-state networks (ESNs) \cite{Jaeger01ESN}, liquid-state machines (LSMs) \cite{Maass02_LSM}, and their unification under the umbrella term of reservoir computing \cite{Verstraeten07RC}, there have been many variations of RCs in terms of mathematical formulation and topology \cite{appeltant2011delay,cuchiero2021sigsas,Carroll19_NetStruct,gallicchio2017deep,Nakajima20phyRC_intro}. 

In this paper we examine the performance of CT, LI, and NG RC setups on tasks requiring multifunctionality. We now outline the method of using each RC for time series prediction.

\subsection{Continuous-time RC}\label{sssec:CTRC}

We use the CT-RC setup introduced in \cite{LuHuntOtt18RC}. During the training stage, the CT-RC is driven by an input signal, $\mathbf{u}(t)$, and its response is described by,
\begin{align}
    \dot{\mathbf{r}}_{CT}(t) = \gamma \left[ - \mathbf{r}_{CT}(t) + \tanh{\left( \textbf{M} \mathbf{r}_{CT}(t) + \sigma \textbf{W}_{in} \mathbf{u}(t) \right)} \right].\label{eq:ListenRes}
\end{align}
$\mathbf{r}_{CT}(t) \in \mathbb{S} \subset \mathbb{R}^{N}$ is the state of the CT-RC in state space, $\mathbb{S}$, at a given time $t$, $N$ is the number of neurons, and $\mathbf{r}_{CT}(0) = \textbf{0}$. $\gamma$ is a time-scale parameter. $\textbf{M} \in \mathbb{R}^{N \times N}$ is the adjacency matrix. The input strength parameter, $\sigma$, and the input matrix, $\textbf{W}_{in} \in \mathbb{R}^{N \times D}$, assign the weight given to the $D$-dimensional input, $\mathbf{u}(t) \in \mathbb{R}^{D}$, as it's projected into the reservoir. $\textbf{M}$ and $\textbf{W}_{in}$ are randomly initialised and design details are provided in the Appendix. The relationship between multifunctionality and the spectral radius, $\rho$, of $\textbf{M}$ is assessed in Sec.\,\ref{sec:Results}.

Solutions of Eq.\,\eqref{eq:ListenRes} are generated using the 4$^{th}$ order Runge-Kutta method with time step $\tau = 0.01$. From $t = t_{listen}$ to $t = t_{train}$, we store the CT-RCs state in $\textbf{X} = [\mathbf{q}(\mathbf{r}_{CT}(t_{listen})) \, \ldots \, \mathbf{q}(\mathbf{r}_{CT}(t_{train}))]$ for $\mathbf{q}(\mathbf{r}(t))=( \mathbf{r}(t) \, \mathbf{r}^{2}(t) )^{T}$, and the input in $\textbf{Y} = [ \mathbf{u}(t_{listen}) \, \ldots \, \mathbf{u}(t_{train}) ]$. In order to train the RC we compute a readout layer, $\psi ( \mathbf{r} (t) ) = \textbf{W}_{out} \, \mathbf{q} ( \mathbf{r}(t) )$ using the following ridge regression formula,
\begin{equation}
    \textbf{W}_{out} = \textbf{Y} \textbf{X}^{T} \left( \textbf{X} \textbf{X}^{T} + \beta \, \textbf{I} \right)^{-1}.\label{eq:RidgeReg}
\end{equation}
$\beta$ is the regularisation parameter, and $\textbf{I}$ is the identity matrix of the appropriate dimension. Note that the RCs response from $t = 0$ until $t = t_{listen}$ is not included in the training in order to remove any dependence the RC has on its initialisation.


After training, the `predicting RC' evolves according to,
\begin{align}
\dot{\hat{\mathbf{r}}}_{CT}(t) = \boldsymbol{\gamma} [ - \hat{\mathbf{r}}_{CT}(t) + &\tanh( \,\, \textbf{M} \, \hat{\mathbf{r}}_{CT}(t) \nonumber \\ & + \sigma \textbf{W}_{in}  \textbf{W}_{out} \, \mathbf{q}(\hat{\mathbf{r}}_{CT}(t)) \,\, ) ], \label{eq:MFPredRes}
\end{align}
and $\hat{\mathbf{r}}_{CT}(0) = \mathbf{r}_{CT}(t_{train})$.

\subsection{Leaky integrator RC}\label{sssec:DTRC}


The LI-RC was introduced in \cite{jaeger2007LeakyInt}, its response to a driving input signal, $\mathbf{u}[i]$, during the training stage is given by,
\begin{align}
    \mathbf{r}_{LI}[i + 1] = (1 - \alpha) \mathbf{r}_{LI}[i] + \alpha \tanh ( \textbf{M} \mathbf{r}_{LI}[i] + \sigma \textbf{W}_{in} \mathbf{u}[i] ),\label{eq:train_LIRC}
\end{align}
where $\mathbf{r}_{LI}[i]$ describes the state of the LI-RC in discrete-time with terms equivalently defined as the CT-RC in Sec.\,\ref{sssec:CTRC}. The difference between the CT-RCs and LI-RCs is the role of the leaky-integrator parameter, $\alpha \in \left[ 0, 1 \right]$. For $\alpha = 1$, the influence of the RCs previous state appears only in $\tanh \left( \cdot \right)$.

The LI-RC is trained with the same approach as the CT-RC for the corresponding solutions between $i_{listen}$ and $i_{train}$. The predicting LI-RC is written as,
\begin{align}
    \hat{\mathbf{r}}_{LI}[i + 1] = (1 - \alpha) \hat{\mathbf{r}}_{LI}[i] + & \alpha \tanh( \,\, \textbf{M} \hat{\mathbf{r}}_{LI}[i] \nonumber \\ &+ \sigma \textbf{W}_{in} \textbf{W}_{out} \mathbf{q}( \hat{\mathbf{r}}_{LI}[i]) \,\, ), \label{eq:pred_LIRC}
\end{align}
and $\hat{\mathbf{r}}_{LI}[0] = \mathbf{r}_{LI}[i_{train}]$.

Note, dividing Eq.\,\eqref{eq:pred_LIRC} by $\tau$ and taking the limit as $\tau \to 0$ gives Eq.\,\eqref{eq:MFPredRes} and implies that $\gamma = \alpha \tau$ and $t = i \tau$. In Sec.\,\ref{sec:Results} we compare the influence that different values of $\alpha$ and $\gamma$ have on the LI and CT RCs. 

\subsection{Next generation RC}\label{sssec:NGRC}




The NG-RC we use was introduced in \cite{gauthier2021ngrc}. In this setup, the input data are transformed with a polynomial multiplication dictionary, $\mathbf{P}^{[O]}$, into a higher dimensional state space consisting of the unique polynomials of orders, $O$. For example, transforming a two-dimensional input data point, $\mathbf{u}[i] = (u_{1}[i], u_{2}[i])^T$, with $\mathbf{P}$ for $O = 1, 2$ is written as,
\begin{equation}
    \mathbf{P^{[1,2]}}(\mathbf{u}[i]) = 
    \left( u_{1}[i] \,\, u_{2}[i] \,\, u_{1}^2[i] \,\, u_{2}^2[i] \,\, u_{1}[i]u_{2}[i] \right)^{T}.
	\label{eq:quadratic}
\end{equation}


A time shift expansion function, $\mathbf{L}_k^s$, of the input data is used in \cite{gauthier2021ngrc} to distinguish the NG-RC from nonlinear vector autoregression algorithms \cite{bollt2021explaining}. The $k$ value defines the number of past data points that the current data point is concatenated with, and the $s$ value denotes how far these points are separated in time. Following the example of a two-dimensional input data point, we write this time shift as,
\begin{align}
    \mathbf{L}^{1}_{2}(\mathbf{u}[i]) = \left( u_{1}[i] \,\, u_{2}[i] \,\, u_{1}[i-1] \,\, u_{2}[i-1] \right)^{T}.
\end{align}
The NG-RCs response to $\mathbf{u}[i]$ during training is written as,
\begin{align}
    \mathbf{r}_{NG}[i+1] = \mathbf{P}^{[O]}(\mathbf{L}^s_k(\mathbf{u}[i])).
\end{align}
Therefore, in this example we write $\mathbf{r}_{NG}[i+1] \in \mathbb{R}^{14}$ as,
\begin{align}
    \left( u_{1}[i] \,\, u_{2}[i] \,\, u_{1}[i-1] \ldots u_{1}[i] u_{2}[i-1] \,\, u_{2}[i] u_{2}[i-1] \right)^{T}.
	\label{eq:quadratictime}
\end{align}

In this setup, the reservoir state vector, $\mathbf{r}_{NG}[i]$, is projected using a readout matrix, $\mathbf{W}_{out}$, to resemble, $\Delta \mathbf{u}[i]=\mathbf{u}[i]-\mathbf{u}[i-1]$, for $i > i_{train}$. $\mathbf{W}_{out}$ is found using Eq.\,\ref{eq:RidgeReg}, with the corresponding $\mathbf{X}$ and $\mathbf{Y}$ constructed as follows.

The input training data $\mathbf{Y} = [ \mathbf{u}[i_{warm}], \ldots, \mathbf{u}[i_{train}] ]$ is transformed to the state matrix $\mathbf{X} =\mathbf{q}\left(\mathbf{P}^{[O]} (\mathbf{L}^{s}_{k}(\mathbf{Y}))\right)$. Note, a warm up time of $i_{warm} = ks$ is needed, where entries of the state matrix at time $i < i_{warm}$ are not defined. The output target matrix used in Eq.\,\ref{eq:RidgeReg} for this setup is written as,
$\mathbf{Y'} = \mathbf{Y}[i] - \mathbf{Y}[i-1]$.
The trained NG-RC evolves according to,
\begin{align}
    \hat{\mathbf{r}}_{NG}[i+1] = \mathbf{P}^{[O]} (\mathbf{L}^{s}_{k}( \hat{\mathbf{u}}_{NG}[i-1] +  \mathbf{W}_{out} \mathbf{q} \left(\hat{\mathbf{r}}_{NG}[i]\right))),
\end{align}
for $\hat{\mathbf{r}}_{NG}[0] = \mathbf{r}_{NG}[i_{train}]$ and $\hat{\mathbf{u}}_{NG}[i-1] = \mathbf{u}[i_{train}] + \sum_{i_{train}}^{i-1} \mathbf{W}_{out} \mathbf{q} \left( \hat{r}_{NG}[i] \right)$ estimates $\mathbf{u}[i]$ for $i > i_{train}$.


As the NG-RC requires tuning only $O$, $k$, $s$, and $\beta$, the issues which relate to improper random initialisations of $\textbf{M}$ and $\textbf{W}_{in}$ do not arise. On the other hand we see in Sec.\,\ref{sec:Results} that there are limitations to what can be achieved with the NG-RC because of its design in comparison to CT and LI-RCs.


\subsection{Training each RC to achieve multifunctionality}\label{ssec:RCtraining}

The same steps are used in training each RC to achieve multifunctionality and are outlined as follows.

For multifunctionality, we require the same $\textbf{W}_{out}$ to hold for $\psi ( \hat{\mathbf{r}}_{\mathcal{S}_{1}} (t) ) \approx \mathbf{u}_{\mathcal{P}_{1}}(t)$ and $\psi ( \hat{\mathbf{r}}_{\mathcal{S}_{2}} (t) ) \approx \mathbf{u}_{\mathcal{P}_{2}}(t)$ for $t > t_{train}$ in the CT-RC and $i > i_{train}$ in the LI and NG-RCs . $\hat{\mathbf{r}}_{\mathcal{S}_{1}}$ and $\hat{\mathbf{r}}_{\mathcal{S}_{2}}$ describe the state of each RC on the coexisting attractors, $\mathcal{S}_{1}$ and $\mathcal{S}_{2}$, which are the RC's representation of the time series described by $\textbf{u}_{\mathcal{P}_{1}}$ and $\textbf{u}_{\mathcal{P}_{2}}$ that each RC is required to reconstruct a coexistence of.

To do this, we generate each RCs response to $\mathbf{u}_{\mathcal{P}_{1}}$ and store it in $\textbf{X}_{\mathcal{S}_{1}}$. The same process is repeated for $\mathbf{u}_{\mathcal{P}_{2}}$ to obtain $\textbf{X}_{\mathcal{S}_{2}}$. These RC training data matrices are concatenated as, $\textbf{X}_{C} = ( \textbf{X}_{\mathcal{S}_{1}}, \, \textbf{X}_{\mathcal{S}_{2}} )$, and similarly for the corresponding $\textbf{Y}_{\mathcal{P}_{1}}$ and $\textbf{Y}_{\mathcal{P}_{2}}$ to obtain $\textbf{Y}_{C}$. $\textbf{W}_{out}$ is calculated using Eq.\,\ref{eq:RidgeReg} for $\textbf{X} = \textbf{X}_{C}$ and $\textbf{Y} = \textbf{Y}_{C}$.

As all RCs are trained using Eq.\,\ref{eq:RidgeReg}, in Sec.\,\ref{sec:Results} we compare the performance of each RC for different values of $\beta$.

\section{Numerical experiments and analysis tools}\label{sec:BenchmarkTests}

In this section we outline the specifics of each numerical experiment used to test the limits of multifunctionality.


\subsection{Coexisting chaotic attractors}\label{ssec:Task1_CAs}

We train each RC to provide a coexistence of the chaotic Lorenz attractor, $\mathcal{L}$, described by,
\begin{align}
    \dot{x} &= 10 ( y - x ), \nonumber\\
    \dot{y} &= x ( 28 - z ) - y, \label{eq:LorenzEq}\\
    \dot{z} &= x y - \frac{8}{3} z + x, \nonumber
\end{align}
and the chaotic Halvorsen attractor, $\mathcal{H}$, described by,
\begin{align}
    \dot{x} &= - 1.3 x - 4 y - 4 z - y^{2}, \nonumber\\
    \dot{y} &= -1.3 y - 4 z - 4 x- z^{2}, \label{eq:HalvorsenEq}\\
    \dot{z} &= -1.3 z - 4 x - 4 y - x^{2}.\nonumber
\end{align}
This pairing of attractors was studied in \cite{herteux2020Symm} when investigating the relationship between symmetry and `mirror attractors'. 

The training data is obtained by generating solutions of Eqs.\,\ref{eq:LorenzEq}-\ref{eq:HalvorsenEq} using the 4$^{th}$ order Runge-Kutta method with time step $\tau=0.01$. Trajectories on $\mathcal{L}$ and $\mathcal{H}$ are normalised such that the furthest point away from the origin on each attractor is less than $1$. In our numerical experiments we move both normalised attractor data sets equidistantly in opposite directions along the z-axis with shift parameter, $\delta z$.

\subsection{The `seeing double' problem}\label{ssec:Task2_SD}

For the second task we consider training each RC to reconstruct trajectories on two completely overlapping circular orbits rotating in opposite directions. We call this paradigmatic multifunctionality task, the `seeing double' problem. 

We generate the respective input sequences for the RC with,
\begin{equation}
\mathbf{u}(t)=
    \left( \begin{array}{c}
        x(t) \\
        y(t)
    \end{array} \right)
    = \left( \begin{array}{c}
        c_{x} \cos{\left( t \right)}\\
        c_{y} \sin{\left( t \right)}
    \end{array} \right).
    \label{eq:InputSys}
\end{equation}
We use Eq.\,\ref{eq:InputSys} to construct a time-series which resembles a trajectory around a circle of radius $c=|c_{x}|=|c_{y}|$ and centered at $\left( 0, 0 \right)$. Two sets, $\mathcal{C}_{A}$ and $\mathcal{C}_{B}$, are produced with Eq.\,\ref{eq:InputSys} and the corresponding input time-series are denoted by $\mathbf{u}_{\mathcal{C}_{A}}$ and $\mathbf{u}_{\mathcal{C}_{B}}$. To create $\mathcal{C}_{A}$ we set $c_{x}=c_{y}=5$ and for $\mathcal{C}_{B}$ we set $c_{x}=-5$ and $c_{y}=5$.




\subsection{Analysis tools}\label{ssec:CommentOnMF}



In Sec.\,\ref{sec:Results} we use the following analysis tools to assess performance of the RCs in the task described in Sec.\,\ref{ssec:Task2_SD}. As the data points describing both circles are effectively the same, the RC needs to have a sufficient memory of its previous state in order to remain on the correct circular orbit. To measure a given RCs memory we make use of the memory capacity metric introduced in \cite{jaeger2002short}.



\subsubsection{Roundness}\label{sssec:Roundness}

To determine whether a given RC achieves multifunctionality, we examine the predictions of the trained RC with an error metric called the `roundness'. For both cycles, we first determine if the prediction of a given cycle is indeed a periodic, and then if it is rotating in the correct direction. Following this we compute the roundness as the difference between the radius of the largest and smallest circle needed to enclose and inscribe the predicted cycle. If the maximum roundness of both roundness values is less than a threshold value $=0.5$ (determined from empirical testing), then we say the RC has achieved multifunctionality.

\subsubsection{Memory capacity}\label{sssec:Memory}



We use the `short-term memory' (STM) \cite{jaeger2002short} to assess if, in this sense, a given RCs memory is critical to achieving multifunctionality.

The STM characterises the capability of a RC to remember inputs from the past. It is measured by training the RC to fit an input signal, $\mu[n]$, at time $n$ to its time shifted signal, $\mu[n - j]$, with each point in $\mu$ chosen from a uniformly random i.i.d of real numbers in $[-1, 1]$. Instead of `closing the loop' after training, the RC is driven with the input signal $\mathbf{\mu}$. The STM is then calculated as the square of the correlation between the reservoir output $\textbf{W}^{j}_{out} r[n]$ and the true values given by $\mu[n - j]$ summed over all $j$,
\begin{align}
    \text{STM} = \sum\nolimits_{j}cor^2 \Big( \mu[n-j], \textbf{W}^{j}_{out} r[n] \Big).
\end{align}
$W^{j}_{out}$ depends on $j$ since for every $j$ the training process needs to be repeated. Note that, $\textbf{M}$, is shared by the RC used during prediction and the RC used to measure the STM while the input and output matrices are different.



\subsubsection{Floquet multipliers}\label{sssec:Floquet}

Given the periodic nature of the seeing double problem, by computing the Floquet multipliers of the RC we can determine whether or not a given configuration of the RC can support the coexistence of both $\mathcal{C}_{A}$ and $\mathcal{C}_{B}$ after the training. The Floquet multipliers are the eigenvalues of the `monodromy matrix', $\boldsymbol{Q}$, which is the solution of,
\begin{align}
    \dot{\boldsymbol{Q}}(t) = \boldsymbol{J}(t) \boldsymbol{Q}(t), \quad \boldsymbol{Q}(0) = \textbf{I},
\end{align}
after one period, $T$, of the RCs response to a given $\textbf{u}(t)$ for one period, $T$, during the training. Here $\boldsymbol{J}(t)$ is the Jacobian matrix of the predicting RC and $\textbf{I}$ is the identity matrix.

For a given driving input signal, if the absolute value of any Floquet multiplier, $\lambda_{i}$, is $> 1$ then the limit cycle in $\mathbb{S}$ is unstable or, if the absolute value of the largest Floquet multiplier, $\lambda_{1}$, is $1$ and all other $\lambda_{i}$'s have absolute value $< 1$ then a given limit cycle is stable.

\section{Results}\label{sec:Results}
The training parameters used to generate all the results shown in this section are given in Tables\,\ref{tab:CTRC_Training_params}-\ref{tab:NGRC_Training_params} in the Appendix.
\subsection{Reconstructing Lorenz and Halvorsen}\label{ssec:Results_LorenzHalvorsen}

\begin{figure*}
    \centering
    \includegraphics[scale=0.39]{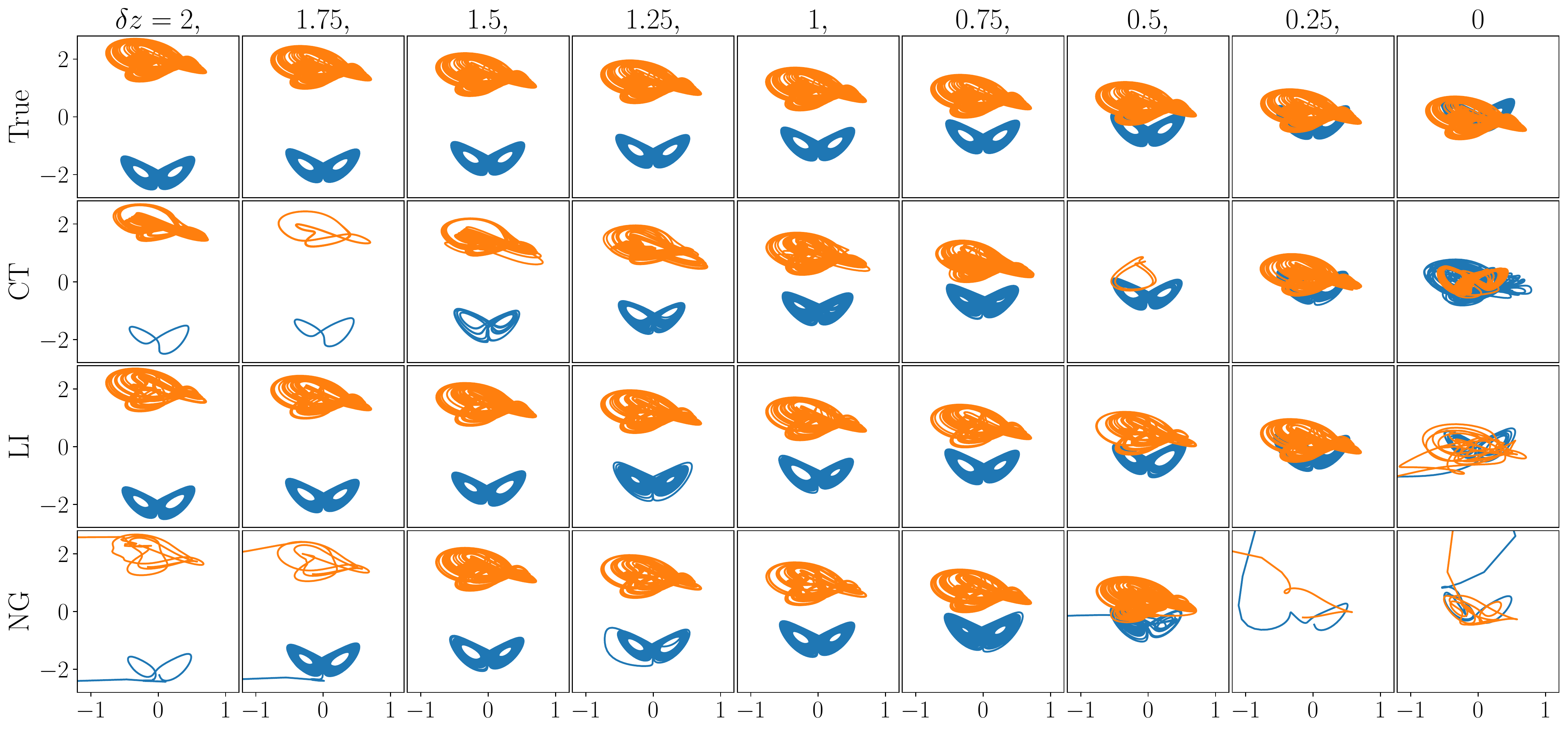}
    \caption{Comparison of CT (continuous-time), LI (leaky-integrator), NG (next-generation) RCs performance when trained to provide of a coexistence of the Lorenz attractor (blue) and the Halvorsen attractor (orange) when separated by a factor of $\delta z$ away from the origin in the $(x, z )$-plane.}
    \label{fig:Lorenz_Halvorsen_RCcomparison}
\end{figure*}

In this section we discuss the results shown in Fig.\,\ref{fig:Lorenz_Halvorsen_RCcomparison} where each RC is trained on the task described in Sec.\,\ref{ssec:Task1_CAs}. 

In Fig.\,\ref{fig:Lorenz_Halvorsen_RCcomparison} we see that all RCs reconstruct a coexistence of $\mathcal{L}$ and $\mathcal{H}$ for $0.75 \leq \delta \leq 1.25$. However, there are some noticeable differences in the performance of each RC as the attractors are brought closer together and further apart. For the set of training parameters specified in the Appendix, no RC is capable of reconstructing a coexistence of $\mathcal{L}$ and $\mathcal{H}$ for all $\delta z$'s and in particular for $\delta z = 0$. The NG-RC is unable to reconstruct both attractors in the case of overlapping data, the CT and LI-RC are able to provide reasonable reconstruction of both $\mathcal{L}$ and $\mathcal{H}$ for $\delta z=0.25$ where large parts of the attractors are already overlapping. This relationship between multifunctionality and overlap is not trivial as, for instance, the CT-RC fails to achieve multifunctionality at times when there is less of an overlap between the attractors , i.e., for $\delta=0.5$ the CT-RC only reconstructs $\mathcal{L}$ but for $\delta=0.25$ both are reconstructed. For the CT-RC and NG-RC we see that as $\mathcal{L}$ and $\mathcal{H}$ are moved away from one another, both RCs fail to achieve multifunctionality with the CT-RCs predictions decaying to limit cycles and the NG-RC becoming unstable. 

As a further comment, in our experiments we found that the CT and NG-RCs achieve multifunctionality for $\delta > 1.5$ with different training parameters. However, by keeping the parameters fixed this provides better insight towards the complex relationship between multifunctionality and overlapping training data. In the modes of failure, the main difference between the NG-RC and the `traditional' CT and LI-RCs is that the state of the NG-RC tends to infinity whereas the CT and LI-RCs prediction decay to some other attractor.

\subsection{Solving the seeing double problem}\label{ssec:Results_SD}

In this section we compare the performance of each RC on the seeing double problem and assess the effects that different training parameters have on multifunctionality.


\subsubsection{LI-RC: $\rho$ and memory capacity}


\begin{figure*}
    \centering
    \begin{subfigure}{0.3\textwidth}
    \centering
    \includegraphics[scale=0.38]{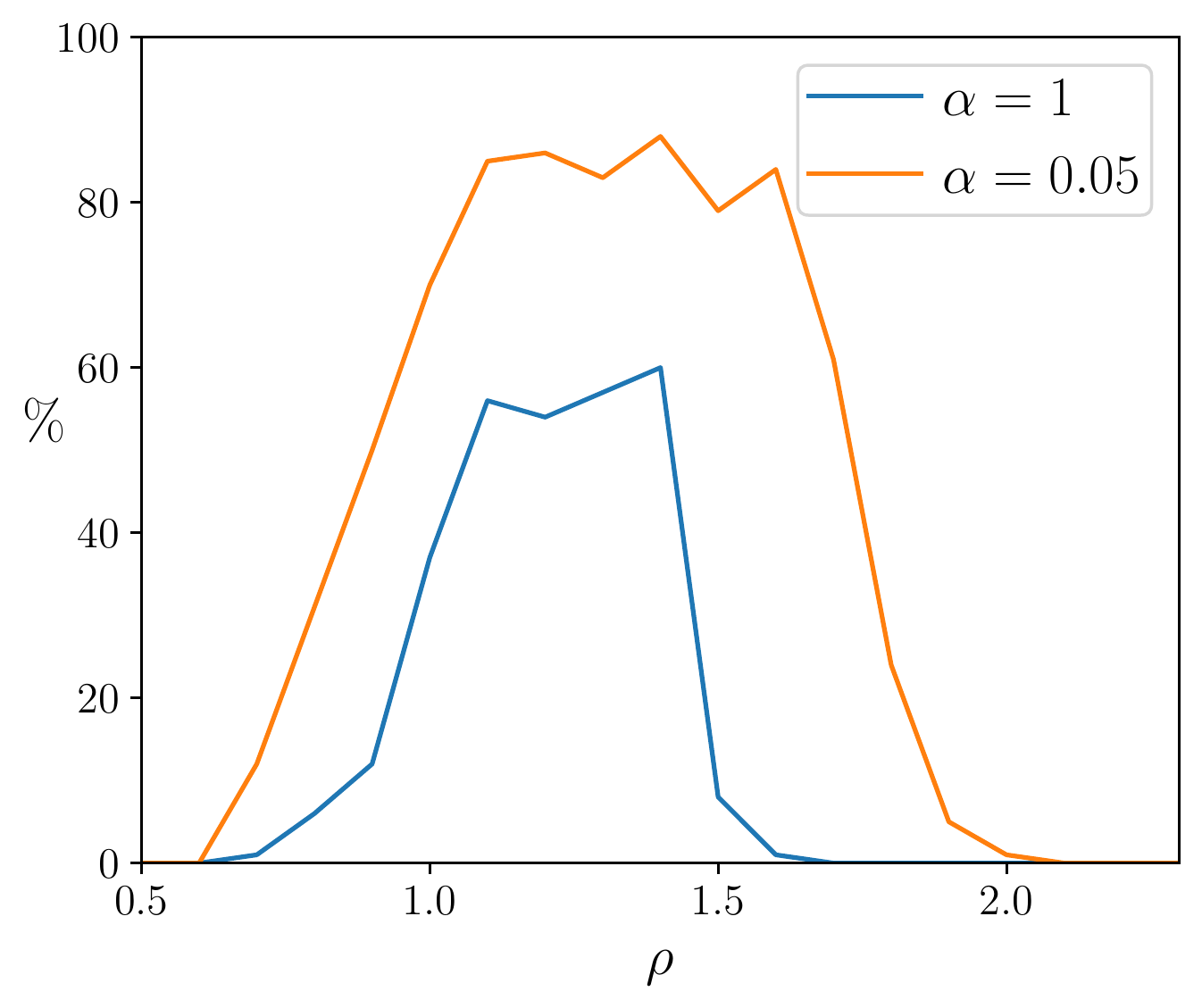}
    \caption{$\rho$ vs. Success}
    \label{fig:100LIRC_rho}
    \end{subfigure}
    \begin{subfigure}{0.22\textwidth}
    \centering
    \includegraphics[scale=0.37]{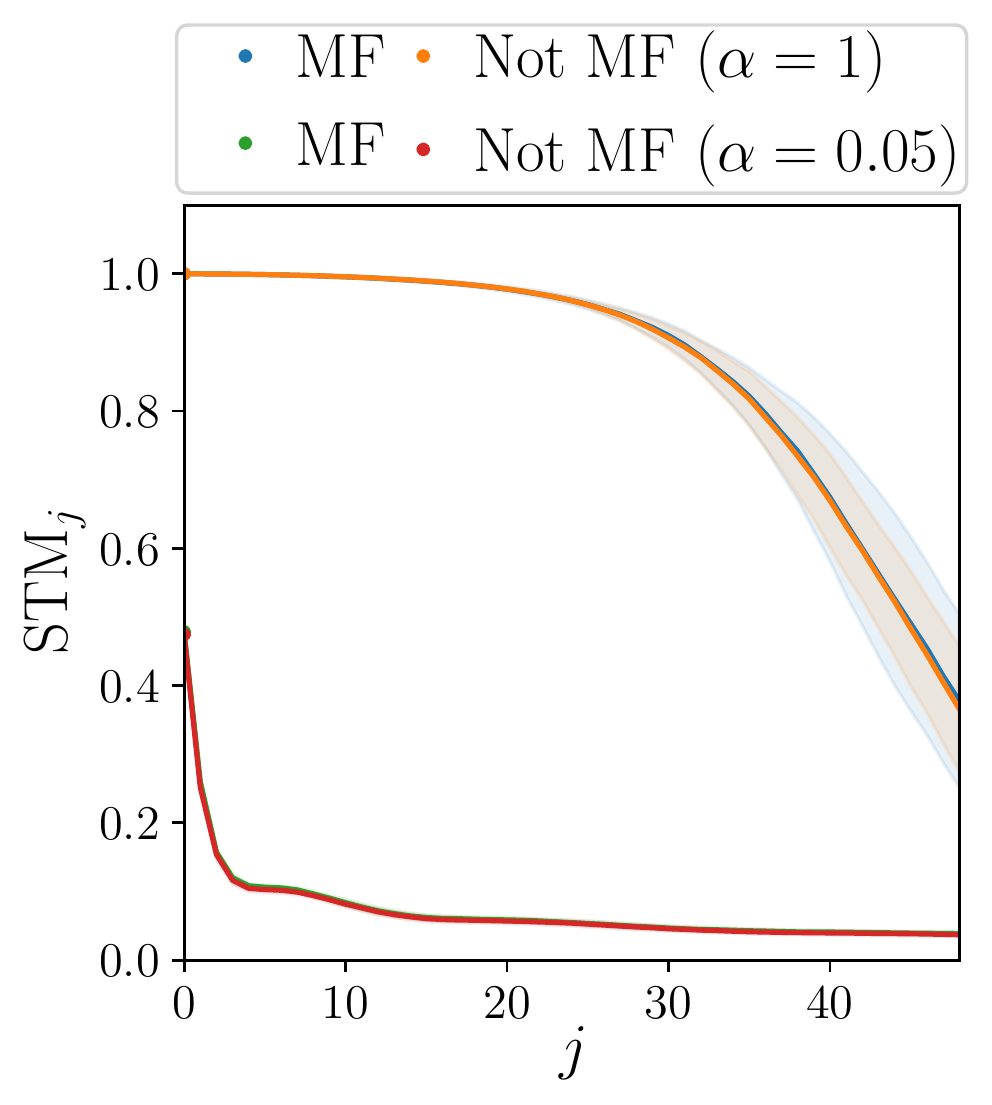}
    \caption{$\rho=1.0$}
    \label{fig:LIRC_Memory_Capacity_rho1}
    \end{subfigure}
    \begin{subfigure}{0.22\textwidth}
    \centering
    \includegraphics[scale=0.37]{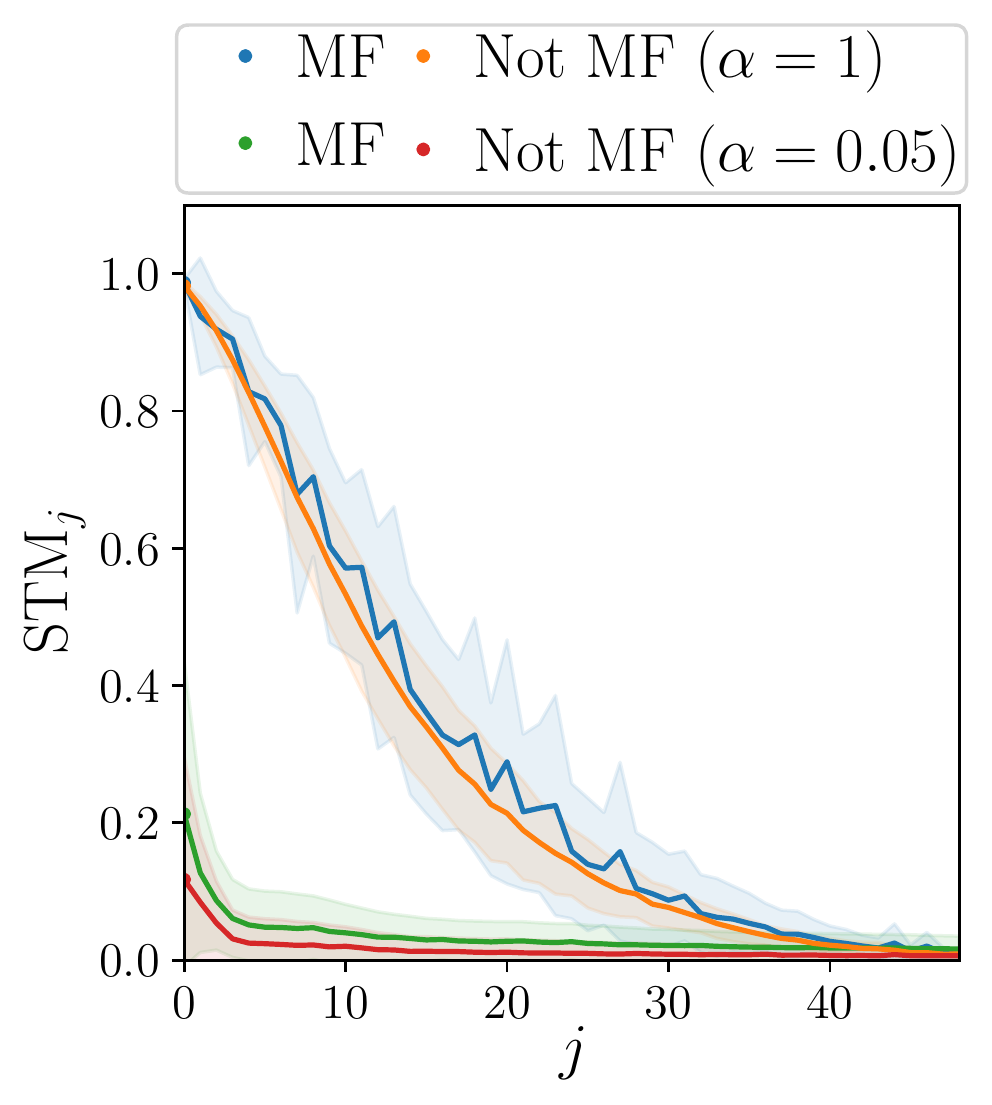}
    \caption{$\rho=1.4$}
    \label{fig:LIRC_Memory_Capacity_rho14}
    \end{subfigure}
    \begin{subfigure}{0.22\textwidth}
    \centering
    \includegraphics[scale=0.37]{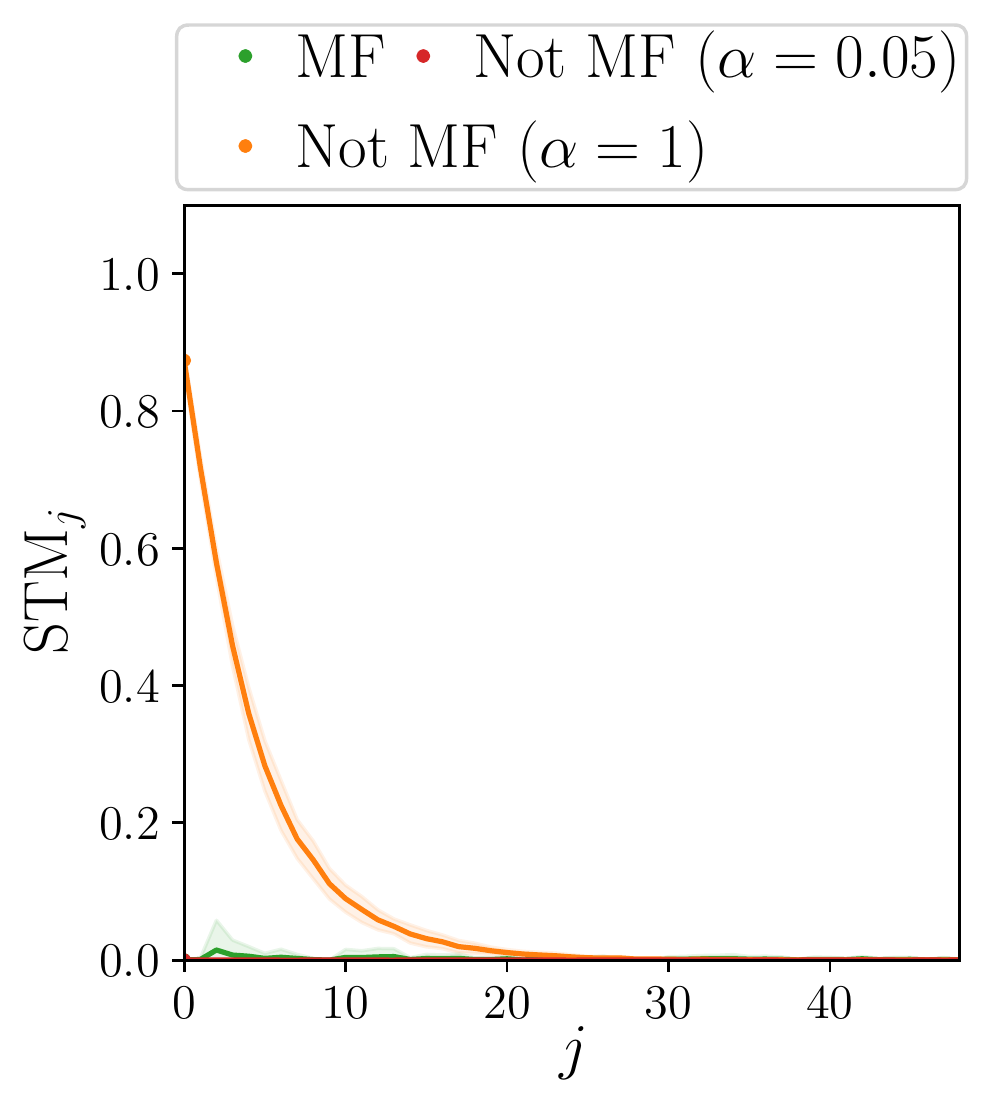}
    \caption{$\rho=1.8$}
    \label{fig:LIRC_Memory_Capacity_rho18}
    \end{subfigure}
    \caption{LI-RC: (a) $\rho$ vs. number of times out of 100 that a given random realisation of $\textbf{M}$ and $\textbf{W}_{in}$ gave rise to multifunctionality. (b)-(d) Short-term memory capacity for RCs that fail and succeed to solve the seeing double problem with error bars given by the standard deviation. $j$ is the time shift used in calculating the STM. MF describes when multifunctionality was achieved.}
    \label{fig:rho_LIRC}
\end{figure*}

In Fig.\,\ref{fig:100LIRC_rho} we illustrate the number of times out of 100 random realisations of $\textbf{M}$ and $\textbf{W}_{in}$ that a given LI-RC gave rise to multifunctionality for different values of $\rho$ as determined by the roundness analysis procedure described in Sec.\,\ref{sssec:Roundness}. We see that for $\alpha = 0.05$ these LI-RCs outperform the LI-RCs with $\alpha = 1$ resulting in not only a broader range of $\rho$ values where multifunctionality was achieved but also a greater probability of success.

In order to determine if the RCs memory as measured with the STM approach discussed in Sec.\,\ref{sssec:Memory} plays an important role in a given RCs ability to achieve multifunctionality, in Figs.\,\ref{fig:LIRC_Memory_Capacity_rho1}-\ref{fig:LIRC_Memory_Capacity_rho18} we compute the STM for ten LI-RCs which achieve multifunctionality and for ten LI-RCs that do not at different $\rho$ values chosen from Fig.\,\ref{fig:100LIRC_rho}. While there is no significant difference in the STM amongst successful or unsuccessful realisations of the LI-RCs, there is a significant difference in the STM for $\alpha=1$ compared to LI-RCs with $\alpha=0.05$. Given that the LI-RC performed much better for $\alpha = 0.05$ as shown in Fig.\,\ref{fig:100LIRC_rho}, we see here in Figs.\,\ref{fig:LIRC_Memory_Capacity_rho1}-\ref{fig:LIRC_Memory_Capacity_rho18} that there is no correlation between multifunctionality and STM.



\subsubsection{CT-RC: $\rho$ and Floquet multipliers}

In this section we examine the behaviour of the CT-RC for different values of $\rho$ and determine that the Floquet multipliers, as discussed in Sec.\,\ref{sssec:Floquet} is a more suitable analysis tool than the STM to distinguish between whether a given realisation of the CT-RC can given rise to multifunctionality in this scenario.

\begin{figure*}
    \centering
    \begin{subfigure}{0.3\textwidth}
    \centering
    \includegraphics[scale=0.38]{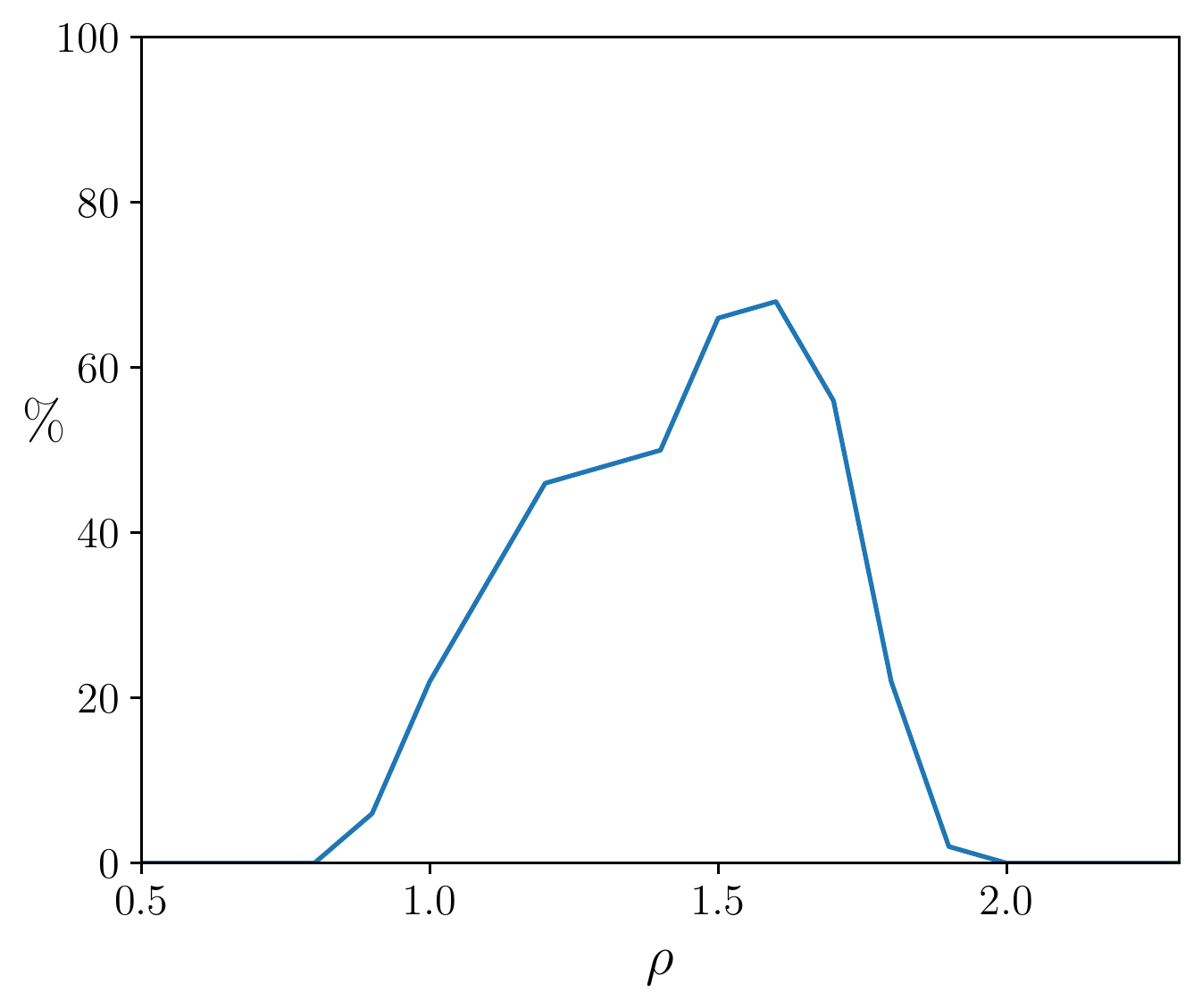}
    \caption{$\rho$ vs. Success}
    \label{fig:100CTRC_rho}
    \end{subfigure}
    \begin{subfigure}{0.22\textwidth}
    \centering
    \includegraphics[scale=0.37]{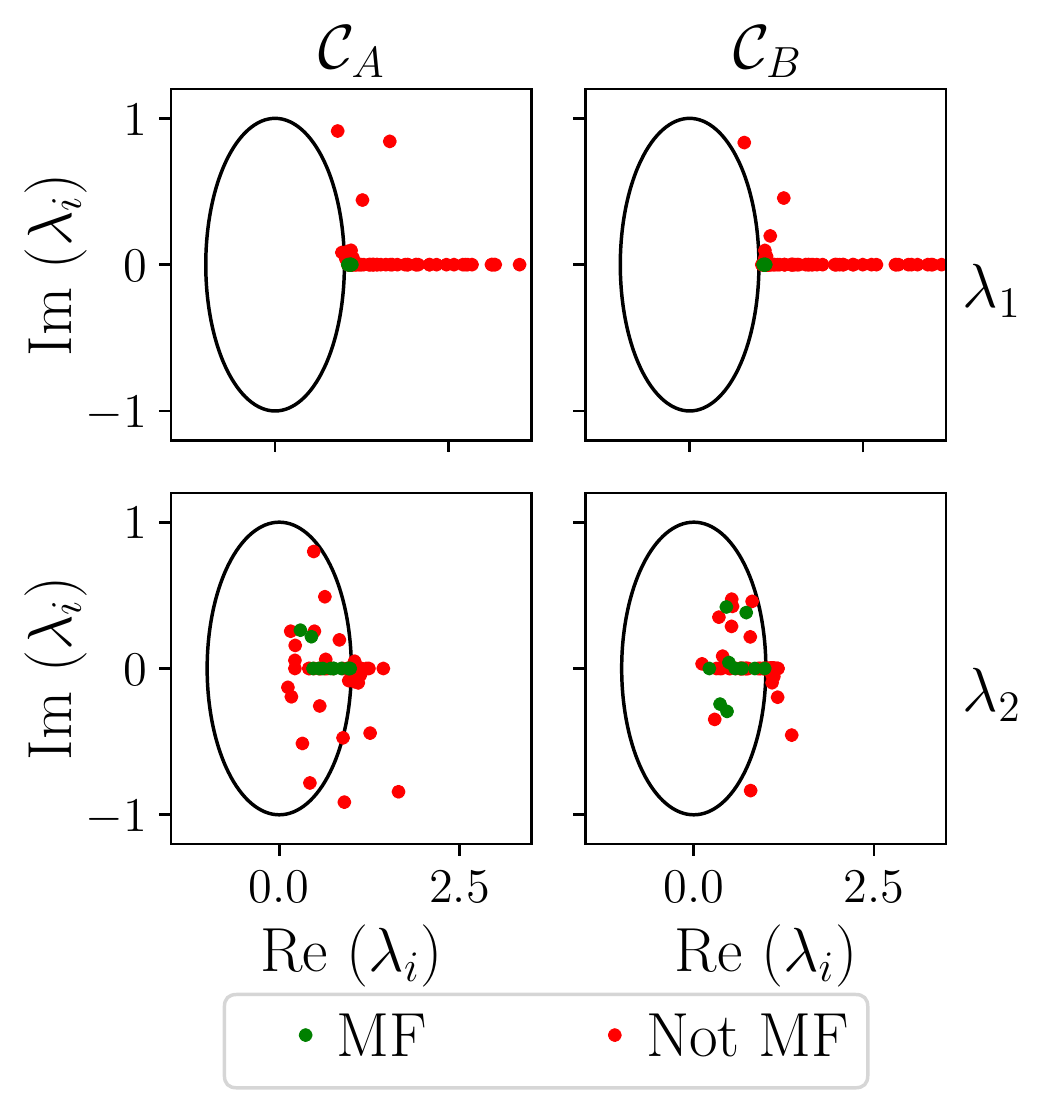}
    \caption{$\rho=1.0$}
    \label{fig:GoodBad_Floquet_2largest_rho1}
    \end{subfigure}
    \begin{subfigure}{0.22\textwidth}
    \centering
    \includegraphics[scale=0.37]{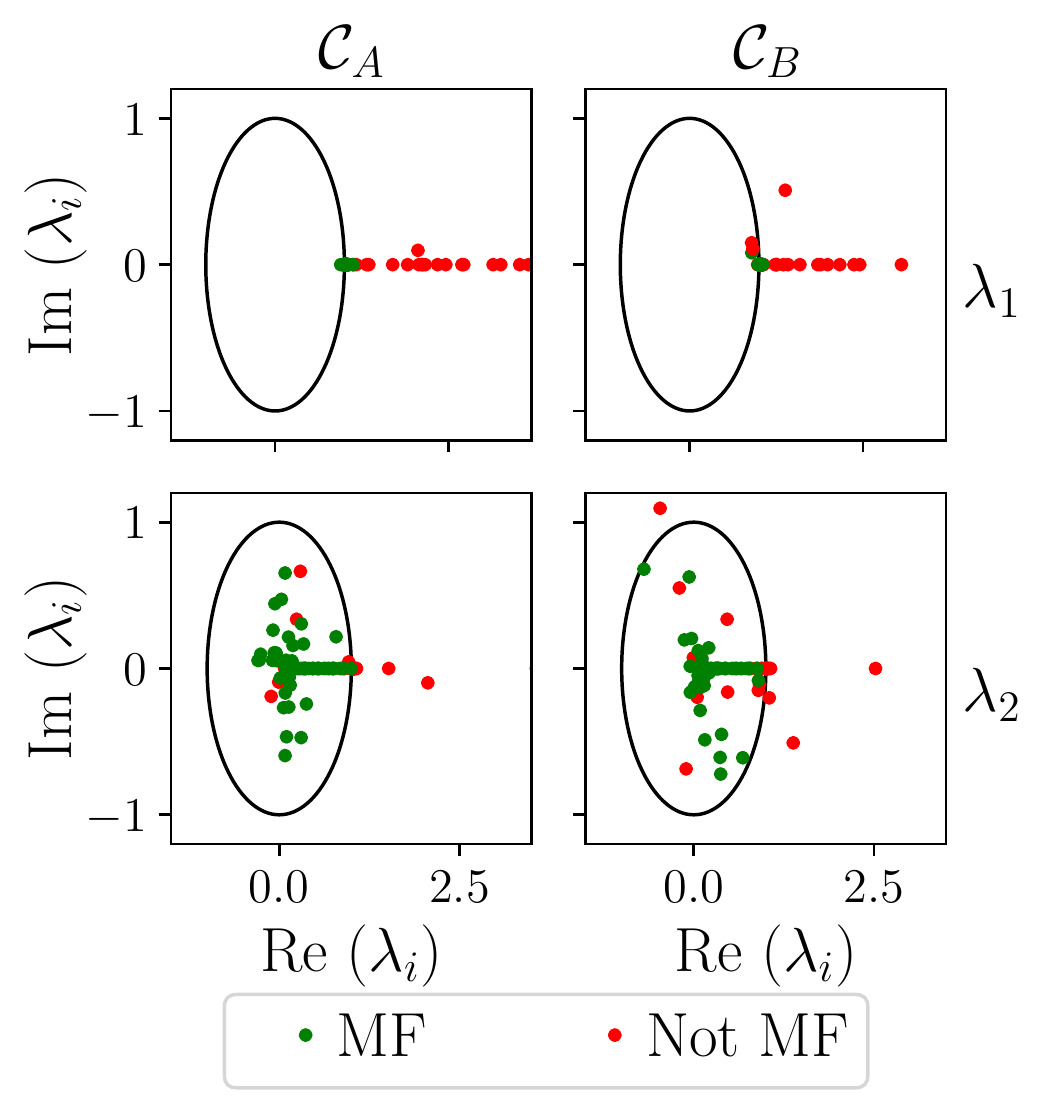}
    \caption{$\rho=1.4$}
    \label{fig:GoodBad_Floquet_2largest_rho14}
    \end{subfigure}
    \begin{subfigure}{0.22\textwidth}
    \centering
    \includegraphics[scale=0.37]{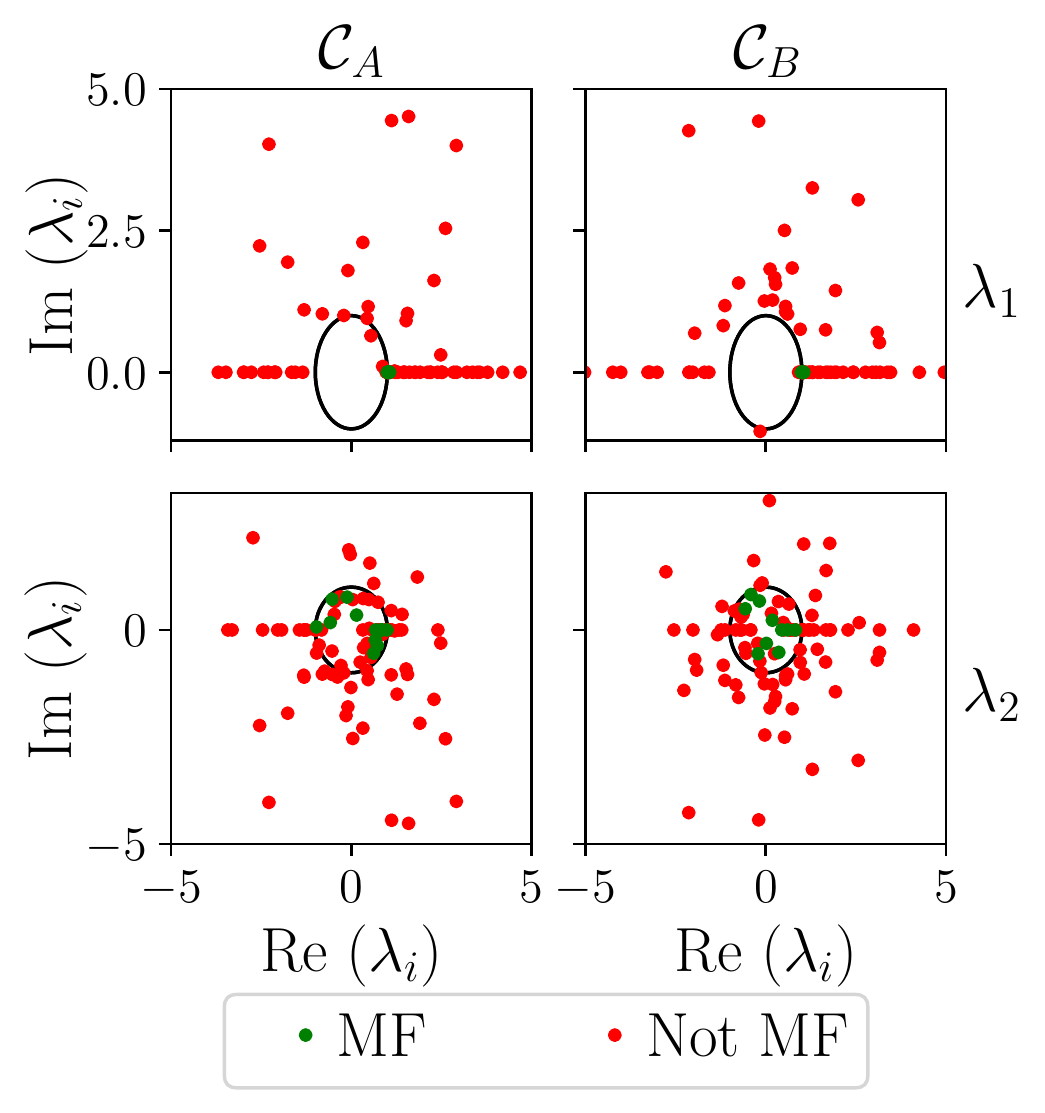}
    \caption{$\rho=1.8$}
    \label{fig:GoodBad_Floquet_2largest_rho18}
    \end{subfigure}
    \caption{CT-RC: (a) $\rho$ vs. number of times out of 100 that a given random realisation of $\textbf{M}$ and $\textbf{W}_{in}$ gave rise to multifunctionality. (b)-(d) Corresponding two largest Floquet multipliers, $\lambda_{1}$ and $\lambda_{2}$, for both $\mathcal{C}_{A}$ and $\mathcal{C}_{B}$ for these $\textbf{M}$ and $\textbf{W}_{in}$ at specified $\rho$. MF describes when multifunctionality was achieved.}
    \label{fig:rho_CTRC}
\end{figure*}

Similarly to Fig.\,\ref{fig:100LIRC_rho}, in Fig.\,\ref{fig:100CTRC_rho} we illustrate the number of times out of 100 that a given random realisation of the CT-RC gave rise to multifunctionality for different values of $\rho$.

In contrast to the LI-RCs with $\alpha = 0.05$, these CT-RCs with $\gamma=5$ are able to achieve multifunctionality with a success ratio $>50\%$ for a smaller range of $\rho$ values but still performs much better than the LI-RCs with $\alpha = 1$. Furthermore, from Figs.\,\ref{fig:GoodBad_Floquet_2largest_rho1}-\ref{fig:GoodBad_Floquet_2largest_rho18} we see that there is a direct correlation between whether a given CT-RC gave rise to multifunctionality and its Floquet multipliers for different values of $\rho$. 

\subsubsection{CT \& LI RCs: $\gamma$, $\alpha$, and $\beta$}

In this section we restrict our analysis to $\rho=1.4$ where we found reasonable performance for both the CT and LI-RCs in order to determine the influence that different values of $\alpha$, $\gamma$, and $\beta$, have on multifunctionality.

\begin{figure}
    \centering
    \begin{subfigure}{0.24\textwidth}
    \centering
    \includegraphics[scale=0.135]{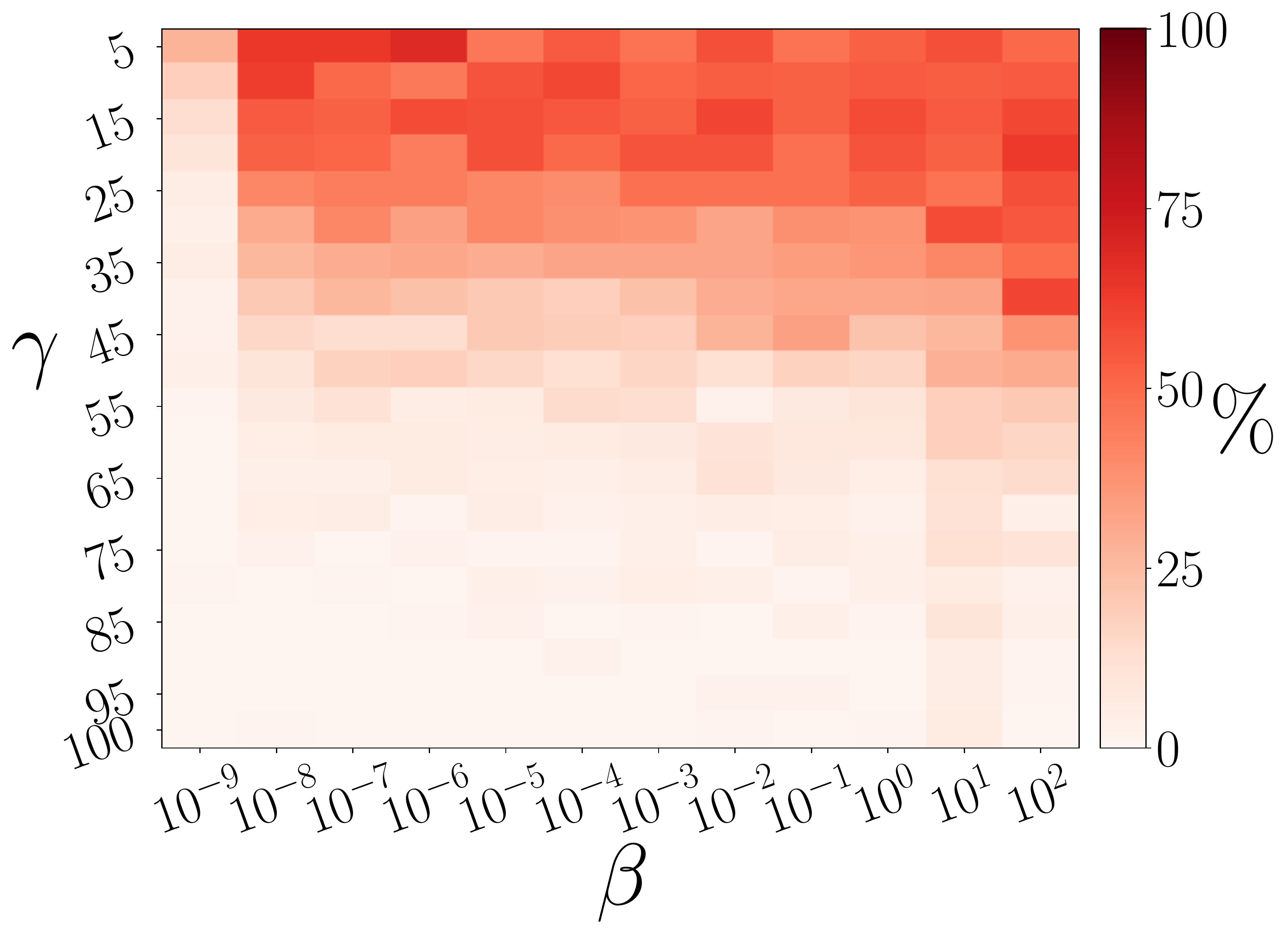}
    \caption{CT-RC}
    \label{fig:CT_beta_m9p2_alpha_10_1_N500_rho14}
    \end{subfigure}
    \begin{subfigure}{0.24\textwidth}
    \centering
    \includegraphics[scale=0.135]{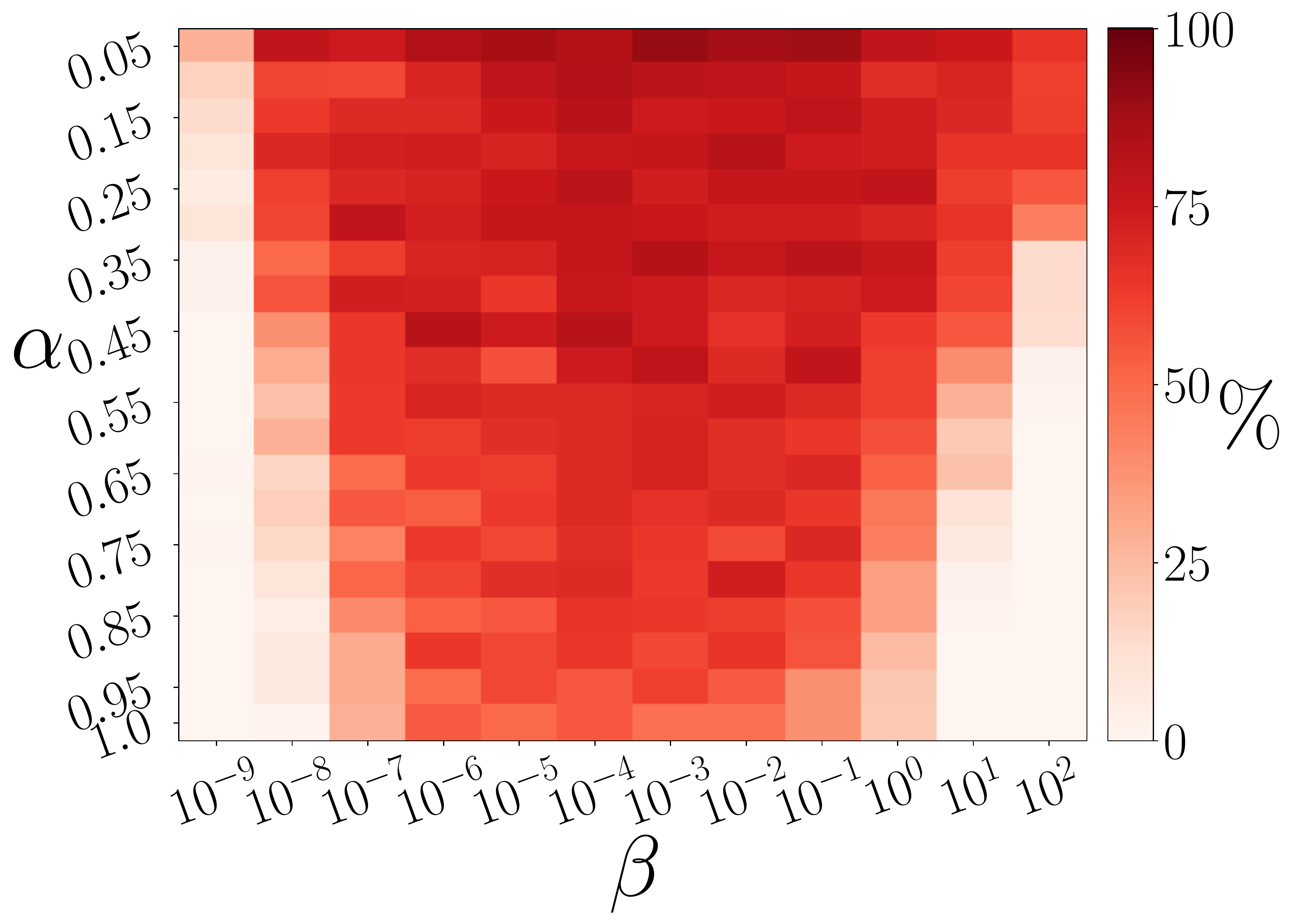}
    \caption{LI-RC}
    \label{fig:LI_beta_m9p2_alpha_10_1_N500_rho14}
    \end{subfigure}
    \caption{Success rate in the $\left( \beta, \gamma \right)$-plane and the $\left( \beta, \alpha \right)$-plane for 100 random realisations of the CT and LI-RCs.}
    \label{fig:alphagammabeta}
\end{figure}

In Fig.\,\ref{fig:alphagammabeta} we plot the number of times out of 100 that a given pair of $( \beta, \alpha )$ or $( \beta, \gamma )$ values give rise to multifunctionality using the error analysis technique as described in Sec.\,\ref{sssec:Roundness}.

In Figs.\,\ref{fig:CT_beta_m9p2_alpha_10_1_N500_rho14} and \ref{fig:LI_beta_m9p2_alpha_10_1_N500_rho14} we see that the limit of $\alpha = \gamma \tau$ for $\tau \to 0$ also holds for only small values of $\alpha$ and $\gamma$. Furthermore, in Fig.\,\ref{fig:CT_beta_m9p2_alpha_10_1_N500_rho14} we see no significant improvement in the performance of the CT-RC for different $\beta$ and $\gamma$ values. In particular, for $\gamma > 40$ the majority of random realisations of the CT-RCs will not give rise to multifunctionality. Whereas for the LI-RC in Fig.\,\ref{fig:LI_beta_m9p2_alpha_10_1_N500_rho14}, its best performance occurs for the same $\alpha \approx 0.05$. However there is a much wider range of $\alpha$ and $\beta$ values where over $70\%$ of the tested LI-RCs achieve multifunctionality. 


What is common between both Figs.\,\ref{fig:CT_beta_m9p2_alpha_10_1_N500_rho14} and \ref{fig:LI_beta_m9p2_alpha_10_1_N500_rho14} is the role of $\beta$. If $\beta$ is too small then both RCs have a success rate $< 5 \%$ and while larger $\beta$ values prevent overfitting, if $\beta$ is too large this also prevents the RCs from learning the correct dynamics. 

\subsubsection{NG-RC}

\begin{figure}
    \centering
    \includegraphics[scale=0.37]{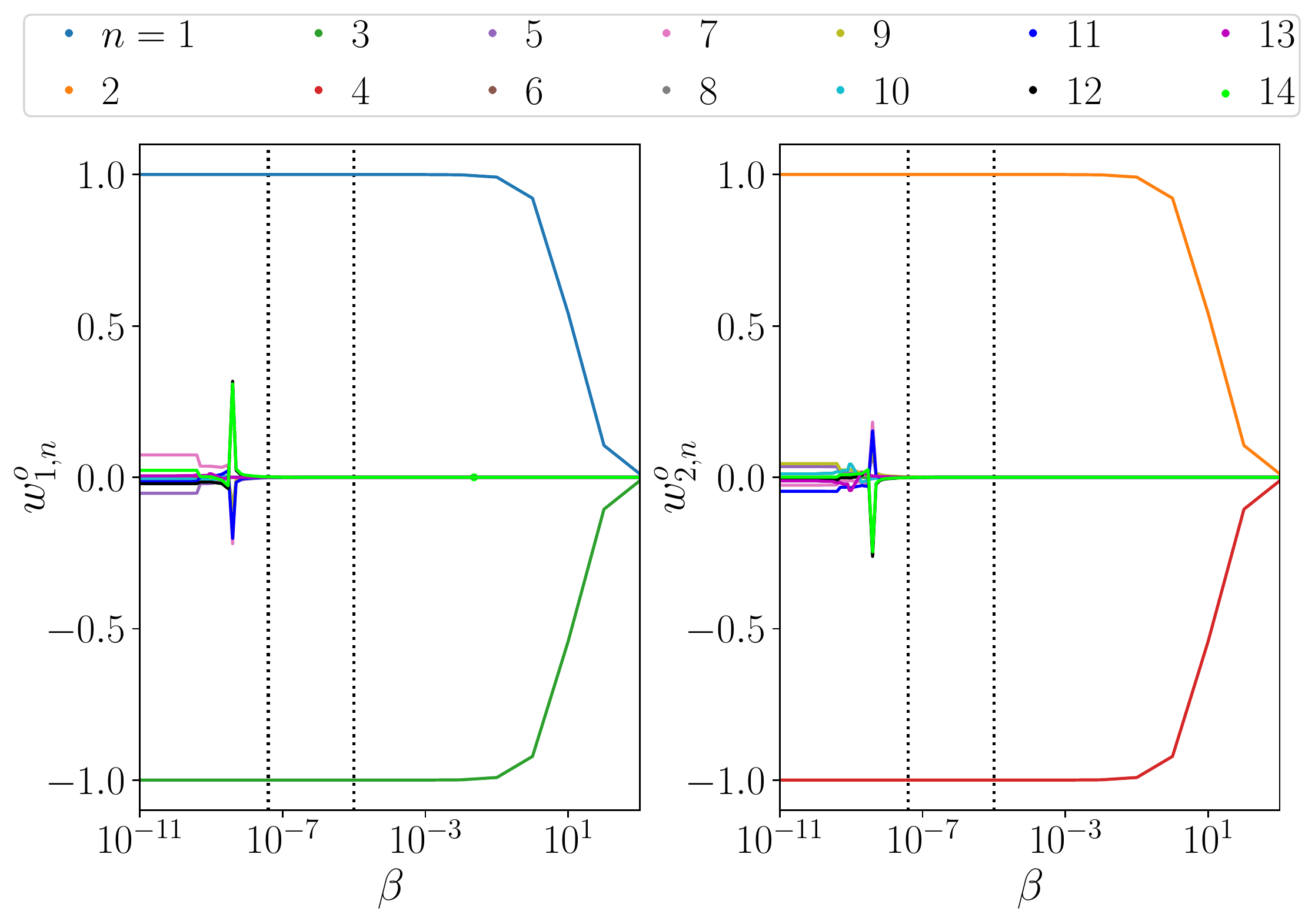}
    \caption{$\beta$ vs. $w_{i,j}^{o}$, the elements of $\textbf{W}_{out}$ for the NG-RC setup}
    \label{fig:beta_NGwout}
\end{figure}

In this section we discuss the results of training the NG-RC on the seeing double task with $O=[1,2], k=2, s=1$. Note the square readout function, $\mathbf{q}(\cdot)$, is not used to obtain $\textbf{W}_{out}$. As a result, the trained $\textbf{W}_{out}$ is $\in \mathbb{R}^{2 \times 14}$, and in Fig.\,\ref{fig:beta_NGwout} we plot each element, $w_{m,n}^{o}$, vs. $\beta$.

As indicated by the vertical dotted lines in Fig.\,\ref{fig:beta_NGwout} we find that the NG-RC only achieves multifunctionality within this range of $\beta$ values. Here there are four $\textbf{W}_{out}$ elements, $w_{1,1}^{o}=0.99989995, w_{1,3}^{o}=-0.99999995, w_{2,2}^{o}=0.99990005$, and $w_{2,4}^{o}=-1.00000005$ which are $\not\approx 0$. For $\beta < 4 \times 10^{8}$, the state of the NG-RC tends to infinity, and for $\beta > 10^{5}$ the system tends to the fixed point at the origin. 

\begin{figure}
    \centering
    \includegraphics[scale=0.35]{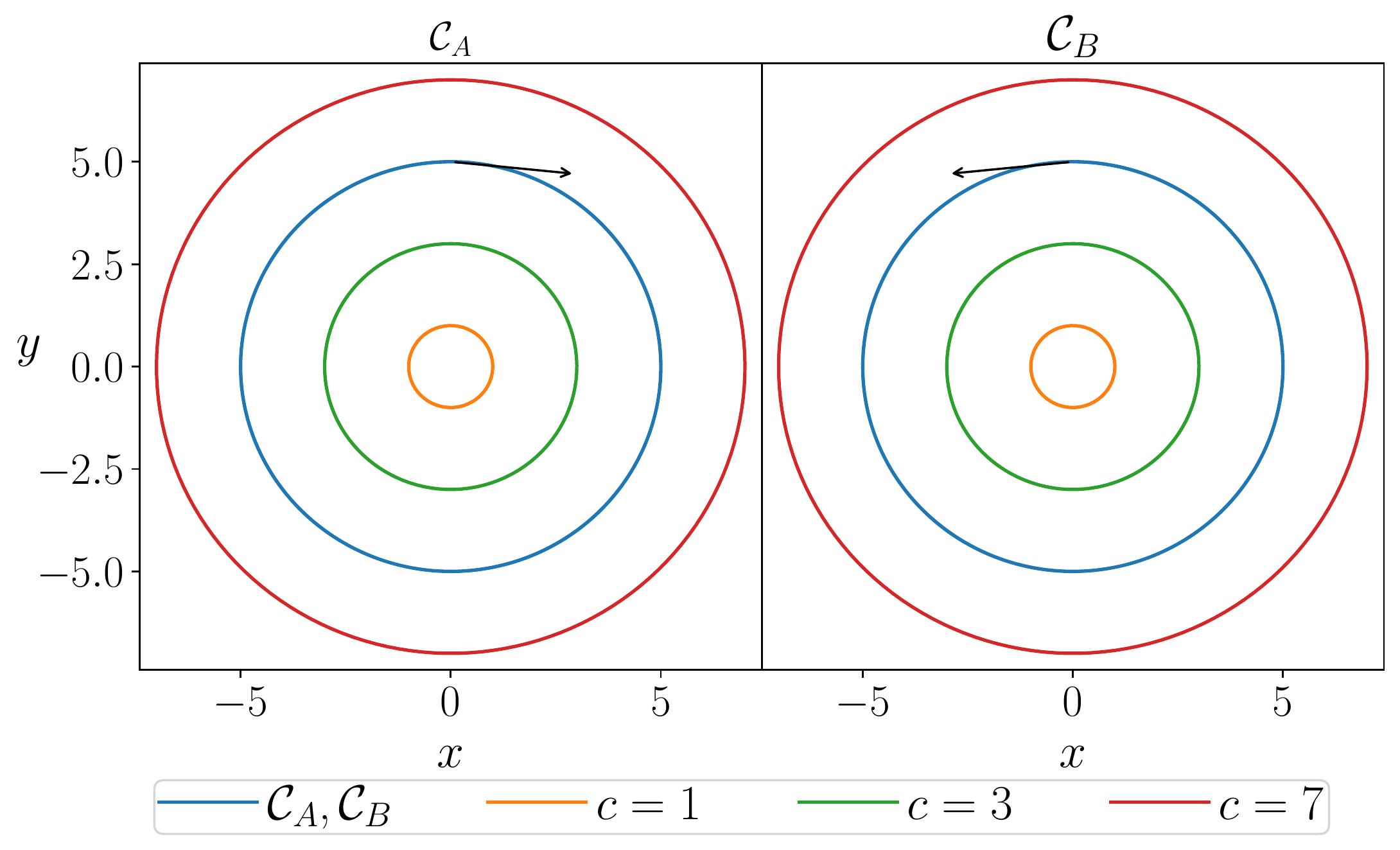}
    \caption{Trained NG-RC output dynamics}
    \label{fig:NGRC_Circles}
\end{figure}

Given the architecture of the NG-RC we are able to write the learned equations as,
\begin{align*}
    x(t+1) & = x(t) + \Delta x(t) = (1 + w_{1,1}^{o}) x(t) + w_{1,3}^{o} x(t-1), \\
    y(t+1) & = y(t) + \Delta y(t) = (1 + w_{2,2}^{o}) y(t) + w_{2,4}^{o} y(t-1).
\end{align*}
Here we see that these governing equations are uncoupled and linear, like the driving input described in Eq.\,\ref{eq:InputSys}. As a consequence, the NG-RC has not reconstructed limit cycles but has produced a set of equations that can mimic the input training data. Furthermore, in Fig.\,\ref{fig:NGRC_Circles} we show that by initialising the trained NG-RC with two consecutive points on a circle it produces circular trajectories of any given radius.





It's also important to note that despite, $w_{1,1}^{o} \approx 1, w_{1,3}^{o} \approx -1, w_{2,2}^{o} \approx 1$, and $w_{2,4}^{o} \approx -1$, the NG-RC becomes unstable if the values $1$ and $-1$ are used for the weights.



\addtolength{\textheight}{-12cm}   

\section{CONCLUSIONS}\label{sec:Conc}

In this paper we investigate some of the limits of multifunctionality in CT, LI, and NG RCs when overlapping training data is introduced in different tasks.

As illustrated in Fig.\,\ref{fig:Lorenz_Halvorsen_RCcomparison}, we find that for a given set of training parameters, each RC can reconstruct a coexistence of the chaotic Lorenz and Halvorsen attractors when the data is sufficiently separated in state space, however if the attractors are too far apart then we see the CT and NG RCs begin to fail. For the same set of training parameters, the NG-RC is unable to reconstruct a coexistence of these attractors once they begin to overlap while the CT and LI RCs are able to achieve multifunctionality when the attractors share mutual regions of state space up to a certain extent.

In order to further explore the limits of multifunctionality, we investigate the performance of each RC when trained to solve the seeing double problem. As shown in Figs.\,\ref{fig:100LIRC_rho} and \ref{fig:100CTRC_rho}, it is clear that $\rho$, the parameter associated with memory in the CT and LI RCs, plays an important role in whether a given RC can achieve multifunctionality. To measure the RCs memory we use the STM metric as described in Sec.\,\ref{sssec:Memory}, and in Figs.\,\ref{fig:LIRC_Memory_Capacity_rho1}-\ref{fig:LIRC_Memory_Capacity_rho18}, we identify the shortcomings of the STM metric as a means to successfully capture the role of memory in this sense. However, by using the Floquet analysis described in Sec.\,\ref{sssec:Floquet} we are able to identify in Figs.\,\ref{fig:GoodBad_Floquet_2largest_rho1}-\ref{fig:GoodBad_Floquet_2largest_rho18} that the effect of `closing the loop' is of greater significance to whether a given RC can achieve multifunctionality. Furthermore, we find that despite choosing a NG-RC with polynomial terms, the trained NG-RC results in a uncoupled set of linear equations which can generate circular trajectories when initialised with two consecutive on any given circle.

The defining characteristic of the reservoir computing approach to machine learning is the need to train only a suitable readout layer, $\textbf{W}_{out}$, to solve a given problem. However, it has only recently been discovered that a given $\textbf{W}_{out}$ can enable a RC to perform more than one task. In this context, other than the results regarding multifunctionality, RCs have been trained to infer unseen attractors, learn global bifurcation structures and anticipate synchronisation \cite{rohm2021unseen,kim2021GlobalLocal,fan2021anticipatingsynch,goldmann2021inferring}.

Multifunctionality opens up new application areas for RCs, for instance, in producing data-driven models of real world phenomenon where multistability is thought to play a role, like in the epileptic brain \cite{suffczynski04EpilepsyDynamics}. However, many questions remain, in particular, how much dynamical functionality a single RC can be trained to exhibit. In future work we aim to address this with further RC designs in other paradigmatic scenarios given the insight gained through the seeing double problem.



\appendix

$\textbf{M} \in \mathbb{R}^{N \times N}$ has a sparse Erd\"{o}s–Renyi network where each element is chosen independently with probability $p$ from a uniform distribution of $(-1,1)$. After initialisation, the spectral radius, $\rho$, of $\textbf{M}$ is tuned in order to provide the network with a sufficient amount of memory. $\textbf{W}_{in} \in \mathbb{R}^{N \times D}$ is designed such that each row has only one nonzero randomly assigned element, chosen uniformly from $(-1,1)$.


\begin{table}[p]
    \centering
    \begin{tabular}{c c c c c c c c c}
    \hline \midrule
        $\text{Fig}$ & $N$ & p & $\rho$ & $\sigma$ & $\gamma$ & $\beta$ & $t_{l}$ & $t_{t}$ \\
        \midrule
        \ref{fig:Lorenz_Halvorsen_RCcomparison} & $1000$ & $0.05$ & $1.6$ & $5$ & $7$ & $10^{2}$ & $100$ & $200$ \\
        \midrule
        \ref{fig:rho_CTRC} & $500$ & $0.05$ & Fig & $0.2$ & $5$ & $10^{-2}$ & $6T$ & $15T$ \\
        \midrule
        \ref{fig:CT_beta_m9p2_alpha_10_1_N500_rho14} & $500$ & $0.05$ & $1.4$ & $0.2$ & Fig & Fig & $6T$ & $15T$ \\
        \midrule \hline\\
    \end{tabular}
    \caption{CT-RC training parameters in the specified figures. $t_{l}=t_{listen}$, $t_{t}=t_{train}$, $T=$ period.}
    \label{tab:CTRC_Training_params}
\end{table}
\begin{table}[p]
    \centering
    \begin{tabular}{c c c c c c c c c}
    \hline \midrule
        $\text{Fig}$ & $N$ & p & $\rho$ & $\sigma$ & $\alpha$ & $\beta$ & $t_{l}$ & $t_{t}$ \\
        \midrule
        \ref{fig:Lorenz_Halvorsen_RCcomparison} & $1000$ & $0.012$ & $0.9$ & $1.2$ & $0.2$ & $10^{-3}$ & $100$ & $200$ \\
        \midrule
        \ref{fig:rho_LIRC} & $500$ & $0.05$ & Fig & $0.2$ & Fig & $10^{-2}$ & $6T$ & $15T$ \\
        \midrule
          \ref{fig:LI_beta_m9p2_alpha_10_1_N500_rho14} & $500$ & $0.05$ & $1.4$ & $0.2$ & Fig & Fig & $6T$ & $15T$ \\
        \midrule
        \hline\\
    \end{tabular}
    \caption{LI-RC training parameters in the specified figures. $t_{l}=t_{listen}$, $t_{t}=t_{train}$, $T=$ period.}
    \label{tab:LIRC_Training_params}
\end{table}
\begin{table}[p]
    \centering
    \begin{tabular}{c c c c c c c}
    \hline \midrule
        $\text{Fig}$ & $O$ & $k$ & $s$ & $\beta$ & $t_{w}$ & $t_{t}$ \\
        \midrule
        \ref{fig:Lorenz_Halvorsen_RCcomparison} & $1,2,3,4,5$ & $3$ & $2$ & $3 \times 10^{-5}$ & $6$ & $200$ \\
        \midrule
        \ref{fig:beta_NGwout}-\ref{fig:NGRC_Circles} & $1,2$ & $2$ & $1$ & Fig & $2$ & $15 T$ \\
        \midrule \hline\\
    \end{tabular}
    \caption{NG-RC training parameters in the specified figures. $t_{w}=t_{warm}$, $t_{t}=t_{train}$, $T=$ period.}
    \label{tab:NGRC_Training_params}
\end{table}



\bibliographystyle{IEEEtran}

\end{document}